\documentclass[letterpaper]{article} 
\usepackage{aaai23}  
\usepackage{times}  
\usepackage{helvet}  
\usepackage{courier}  
\usepackage[hyphens]{url}  
\usepackage{graphicx} 
\usepackage{natbib}  
\usepackage{caption} 
\usepackage{multirow}
\usepackage{xcolor}
\usepackage{subcaption}
\usepackage{booktabs}
\usepackage{amsfonts}
\usepackage{amsmath}
\usepackage{lineno}
\usepackage{pgfplots}

\usepackage[ruled]{algorithm2e}

\usepackage{pgfkeys}
\newenvironment{customlegend}[1][]{%
    \begingroup
    \csname pgfplots@init@cleared@structures\endcsname
    \pgfplotsset{#1}%
}{%
    \csname pgfplots@createlegend\endcsname
    \endgroup
}%
\def\addlegendimage{\csname pgfplots@addlegendimage\endcsname}

\title{Continual Graph Convolutional Network for Text Classification}
\author {
    Tiandeng Wu\textsuperscript{\rm 1}$^{*}$,
    Qijiong Liu\textsuperscript{\rm 2}\thanks{Equal contribution (co-first authors). Author order determined by dice rolling.\\ \indent \hspace{0.1cm}$^{\dag}$Corresponding authors.},
    Yi Cao\textsuperscript{\rm 1},
    Yao Huang\textsuperscript{\rm 1},
    Xiao-Ming Wu\textsuperscript{\rm 2}$^{\dag}$,
    Jiandong Ding\textsuperscript{\rm 1}$^{\dag}$
}
\affiliations {
    \textsuperscript{\rm 1} Huawei Technologies Co., Ltd., China\\
    \textsuperscript{\rm 2} The Hong Kong Polytechnic University, Hong Kong\\
    \{wutiandeng1, caoyi23, huangyao11, dingjiandong2\}@huawei.com, \\
    jyonn.liu@connect.polyu.hk, xiao-ming.wu@polyu.edu.hk
}

\begin{document}

\maketitle

\begin{abstract}
Graph convolutional network (GCN) has been successfully applied to capture global non-consecutive and long-distance semantic information for text classification. However, while GCN-based methods have shown promising results in offline evaluations, they commonly follow a \emph{seen-token-seen-document} paradigm by constructing a fixed document-token graph and cannot make inferences on new documents. It is a challenge to deploy them in online systems to infer steaming text data. In this work, we present a continual GCN model (ContGCN) to generalize inferences from observed documents to unobserved documents. Concretely, we propose a new \textit{all-token-any-document} paradigm to dynamically update the document-token graph in every batch during both the training and testing phases of an online system. Moreover, we design an occurrence memory module and a self-supervised contrastive learning objective to update ContGCN in a label-free manner. A 3-month A/B test on Huawei public opinion analysis system shows ContGCN achieves 8.86\% performance gain compared with state-of-the-art methods. Offline experiments on five public datasets also show ContGCN can improve inference quality. The source code will be released at {\url{https://github.com/Jyonn/ContGCN}}.

\end{abstract}

\section{Introduction}

As one of the fundamental tasks in natural language processing, text classification has been extensively studied for decades and used in various applications~\cite{xu2019sentiment,abaho-etal-2021-detect}. In recent years, graph convolutional network (GCN) has been successfully applied in text classification~\cite{yao2019graph,bertgcn} to capture global non-consecutive and long-distance semantic information such as token co-occurrence in a corpus. 

A line of GCN-based methods~\cite{li2019label} perform document classification by simply constructing a homogeneous graph with each document as a node and modeling inter-document relations such as citation links, which however does not exploit document-token semantic information.
Another line of GCN-based methods constructs heterogeneous document-token graphs, where each {node} represents a document or a token, and each {edge} indicates a correlation factor between two nodes. 
However, they commonly follow a \textit{seen-token-seen-document} (\textbf{STSD}) paradigm to construct a \emph{fixed} document-token graph with all seen documents (labeled or unlabeled) and tokens and perform transductive inference. 
Given a new document with unobserved tokens, the trained model cannot make an inference because neither the document nor the unseen tokens are included in the graph. Hence, while these methods are effective in offline evaluations, they cannot be deployed in online systems to infer streaming text data.

To address this challenge, in this paper, we propose a new \textit{all-token-any-document} (\textbf{ATAD}) paradigm to dynamically construct a document-token graph, and based on which we present a {cont}inual {GCN} model (\textbf{ContGCN}) for text classification. Specifically, we take the vocabulary of a pretrained language model (PLM) such as BERT~\cite{bert} as the {global} token set, so a new document can be tokenized into seen tokens from the vocabulary. 
We further form a document set which may contain any present documents (e.g., those in the current batch). The document-token graph then consists of tokens in the global token set and documents in the document set. The edge weights of the graph are dynamically calculated according to an occurrence memory module with historical token correlation information, and document embeddings are generated with pretrained semantic knowledge. In this way, ContGCN is enabled to perform inductive inference on streaming text data.
Furthermore, to address data distribution shift~\cite{luo2022online} which is prevalent in online services, we design a label-free online updating mechanism for ContGCN to save the cost and effort for periodical offline updates of the model with new text data. Specifically, we fine-tune the occurrence memory module according to the distribution shift of streaming text data and update the network parameters with a self-supervised contrastive learning objective.

ContGCN achieves favorable performance in both online and offline evaluations thanks to the proposed ATAD paradigm and label-free online updating mechanism. We have deployed ContGCN in an online text classification system -- Huawei public opinion analysis system, which processes thousands of textual comments daily. A 3-month A/B test shows ContGCN achieves 8.86\% performance gain compared with state-of-the-art methods. Offline evaluations on five real-world public datasets 
also demonstrate the effectiveness of ContGCN. 

To summarize, our contributions are listed as follows:
\begin{itemize}
    
    \item We propose a novel all-token-any-document paradigm and a continual GCN model to infer unobserved streaming text data, which, to our knowledge, is the first attempt to use GCN for online text classification.
    
    \item We design a label-free updating mechanism based on an occurrence memory module and a self-supervised contrastive learning objective, which enables to update ContGCN online with unlabeled documents.
    
    \item Extensive online A/B tests on an industrial text classification system and offline evaluations on five real-world datasets demonstrate the effectiveness of ContGCN.
\end{itemize}

\section{Preliminaries}

\newcommand{\nNode}{{n}}
\newcommand{\nDim}{{d}}
\newcommand{\nLayer}{{h}}
\newcommand{\nDoc}{{m}}
\newcommand{\nVoc}{{u}}
\newcommand{\am}{\mathbf{A}}
\newcommand{\dm}{\mathbf{D}}
\newcommand{\hm}{\mathbf{H}}
\newcommand{\ssp}{{(\doc)}}
\newcommand{\sj}{{(\mathbf{s}_j)}}
\newcommand{\nm}{\mathbf{X}}
\newcommand{\doc}{\mathbf{s}}
\newcommand{\mm}[2]{\in \mathbb{R}^{#1 \times #2}}
\newcommand{\mmDim}[1]{\mm{#1}{\nDim}}
\newcommand{\mmSqr}[1]{\mm{#1}{#1}}

\subsection{Graph Convolutional Network (GCN)}

GCN~\cite{gcn} is a graph encoder that aggregates information from node neighborhoods. It is composed of a stack of graph convolutional layers.
Formally, we use $\mathcal{G}=(\mathcal{V}, \mathcal{E})$ to denote a graph, where $\mathcal{V}(\nNode=|\mathcal{V}|)$ and $\mathcal{E}$ are sets of nodes and edges, respectively.
Note that each node $v \in \mathcal{V}$ is self-connected, i.e., $(v, v) \in \mathcal{E}$. 
We use $\nm \mmDim{\nNode}$ to represent initial node representations, where $\nDim$ is the embedding dimension. To aggregate information from neighborhoods, a symmetric adjacency matrix $\am \mmSqr{\nNode}$ is introduced, where $\am_{ij}$ is the correlation score of nodes $v_i$ and $v_j$ and $\am_{ii} = 1$. 
The adjacency matrix is normalized as 
\begin{gather}
    \widetilde{\am} = \dm^{-\frac{1}{2}}\am\dm^{\frac{1}{2}},
\end{gather}
where $\dm$ is a degree matrix and $D_{ii} = \sum_{j} A_{ij}$.
At the $k$-th 
convolutional layers, the node embeddings are calculated as:
\begin{gather}
	\hm^{(k)} = \sigma \left( \widetilde{\am}\hm^{(k-1)}\mathbf{W}_{k} \right),
\end{gather}
where $k \in \{1, 2, \ldots, \nLayer\}$, $h$ is the total number of convolutional layers, $\sigma$ is the activation function, and $\mathbf{W}_{k} \mmSqr{\nDim}$ is a trainable matrix. Specifically, $\hm^{(0)} = \nm$.

\subsection{GCN-Based Text Classification  \label{sec:lotgod-preliminary}}

Text classification aims to classify documents into different categories. 
Formally, we use $\mathcal{D}(\nDoc=|\mathcal{D}|)$ to denote a set of given documents, which can be split into a training set $\mathcal{D}_\text{train}$ and a testing set $\mathcal{D}_\text{test}$.
Each document $\doc$ can be represented as a list of tokens, $\doc = (t_1^\ssp, t_2^\ssp, \cdots, t_{|\doc|}^\ssp)$, where $t_i^\ssp \in \mathcal{T}$ is a token in the token vocabulary set $\mathcal{T}(\nVoc = |\mathcal{T}|)$.

\begin{figure}
\centering
\begin{tabular}{cc}
\begin{subfigure}{0.2\textwidth}
    \includegraphics[width=1\textwidth]{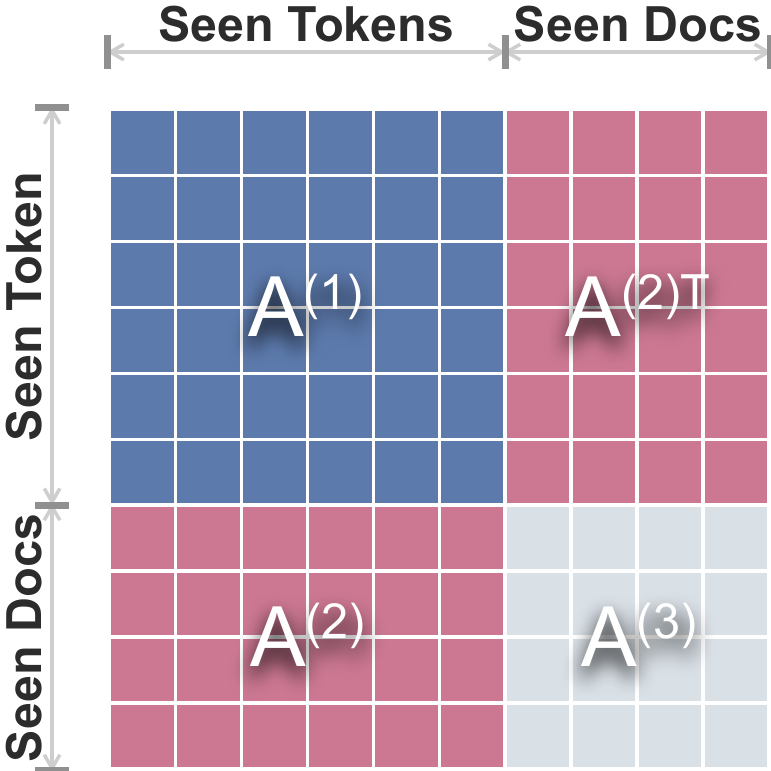}
    \caption{\label{fig:STSD}STSD}
\end{subfigure} & 
\begin{subfigure}{0.2\textwidth}
    \includegraphics[width=1\textwidth]{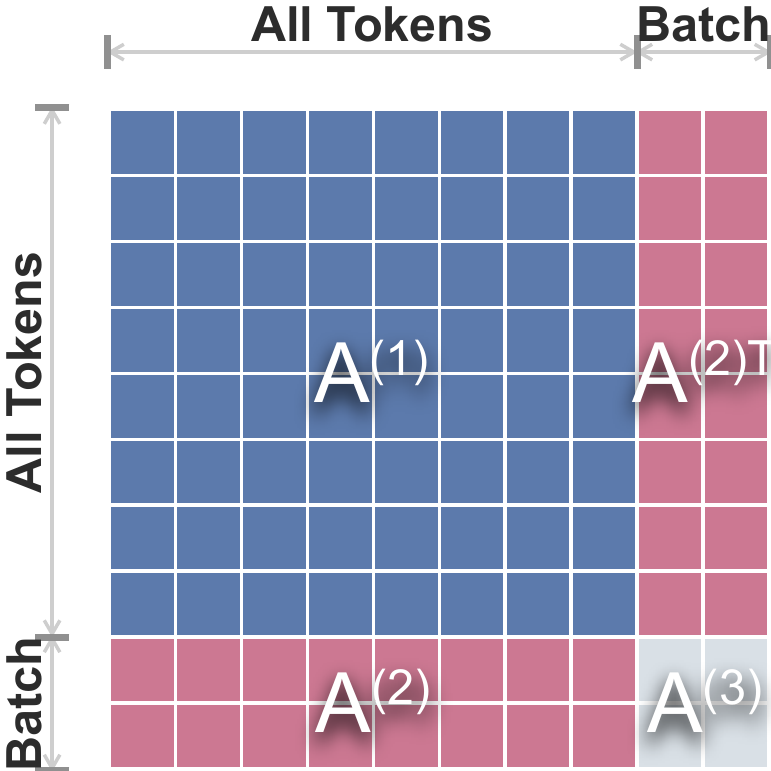}
    \caption{\label{fig:ATAD}ATAD}
\end{subfigure}
\end{tabular}

\caption{\label{fig:STSD-vs-ATAD} Comparison of the adjacency matrices. Left: seen-token-seen-document (STSD) paradigm (e.g., BertGCN). Right: proposed all-token-any-document (ATAD) paradigm.}
\end{figure}

Existing GCN-based text classification methods~\cite{yao2019graph,qiao2018a,bertgcn} mainly follow a seen-token-seen-document paradigm to construct heterogeneous document-token graphs.
Specifically, they first form a seen vocabulary set $\mathcal{T}_\text{seen}$ of size $\nVoc^\prime$ ($\mathcal{T}_\text{seen} \subset \mathcal{T}$), which contains all the seen tokens in the document set $\mathcal{D}$.
Then, they construct a fixed document-token graph, whose nodes include all the seen tokens in $\mathcal{T}_\text{seen}$ and all the given documents in $\mathcal{D}$.
The adjacency matrix $\am$ of the graph is shown in Figure~\ref{fig:STSD}, which consists of a token-token symmetric matrix $\am^{(1)} \mmSqr{\nVoc^\prime}$, a document-token matrix $\am^{(2)} \mm{\nDoc}{\nVoc^\prime}$, and a document-document identity matrix $\am^{(3)} \mmSqr{\nDoc}$. 
The adjacency matrix $\am$ is fixed. Finally, GCN is applied to encode and classify the documents.


\begin{figure*}[htpb]
\centering
\includegraphics[width=0.9\linewidth]{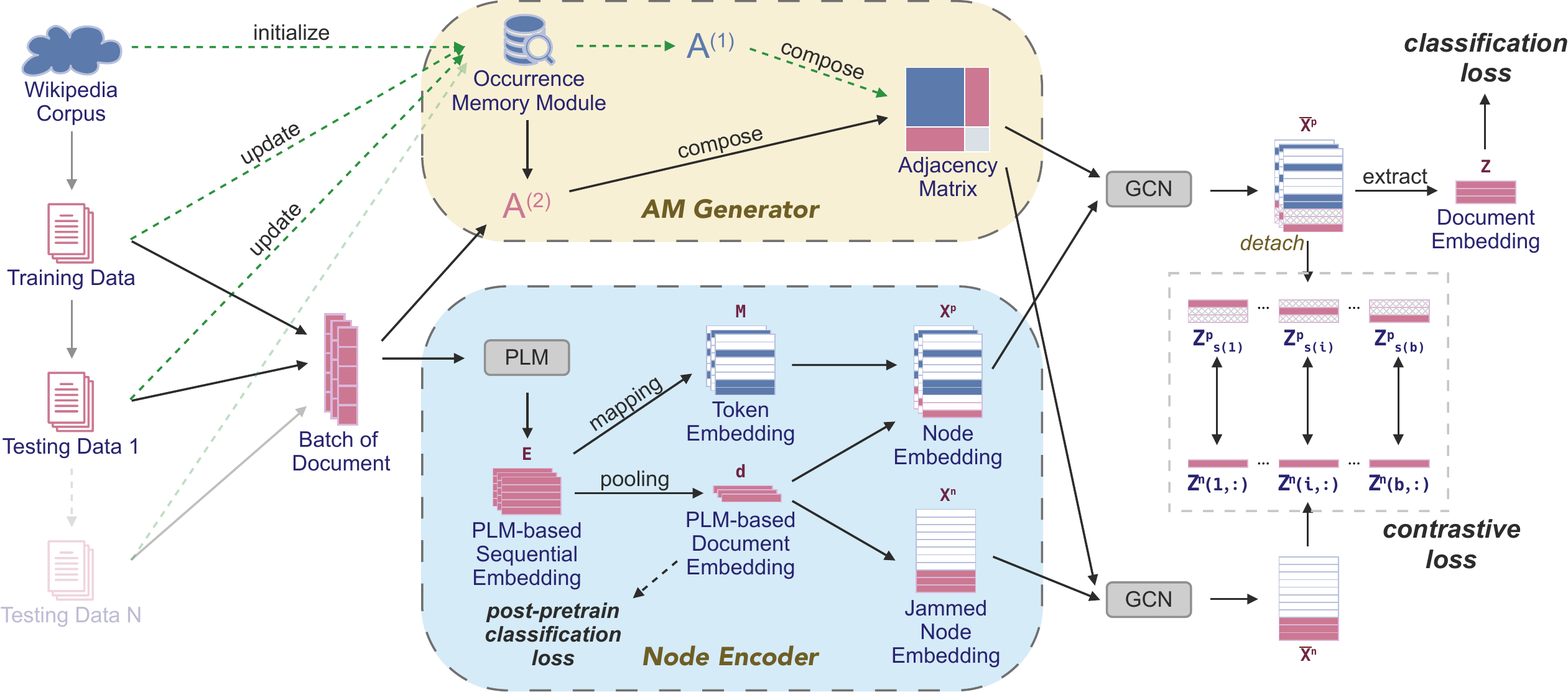}
\caption{Framework of our ContGCN model. Green dotted lines represent operations before each phase of model training or testing. Two key components, i.e., AM Generator and Node Encoder, dynamically construct the adjacency matrix and generate node embeddings, which are then fed into a GCN encoder. Finally, our ContGCN model is trained with a classification loss and an anti-interference contrastive loss. }
\label{fig:overview}
\end{figure*}

\newcommand{\nBat}{{b}}
\newcommand{\nSen}{{s}}
\newcommand{\nLen}{{l}}
\newcommand{\nCls}{{c}}

\section{Method}



The commonly adopted seen-token-seen-document (STSD) paradigm only allows to infer seen documents due to its transductive nature. 
To address this issue, we propose a novel all-token-any-document (ATAD) paradigm that leverages the global token vocabulary and dynamically updates the document-token graph to make inferences on unobserved documents. 

Based on the ATAD paradigm, we propose a continual GCN model, namely ContGCN, as 
illustrated in Figure~\ref{fig:overview}, which comprises of an adjacency matrix generator, a node encoder, and GCN encoders. 
Specifically, given a batch of input documents, the adjacency matrix generator updates the adjacency matrix based on the occurrence memory module and the current batch of documents, while the node encoder produces content-based node embeddings. The GCN encoder is then employed to capture the global-aware node representations. Finally, two training objectives are applied to train the ContGCN model.

\subsection{Proposed All-Token-Any-Document Paradigm}

In contrast to the seen-token-seen-document paradigm, which treats seen tokens ($\mathcal{T}_\text{seen}$) and seen documents ($\mathcal{D}$) as fixed graph nodes, our proposed all-token-any-document paradigm involves constructing a document-token graph with all tokens ($\mathcal{T}$) and dynamic documents (i.e., a batch of documents $B = \{\mathbf{s}_1, \mathbf{s}_2, \cdots, \mathbf{s}_\nBat\} \subset \mathcal{D}$ where $\nBat$ is the batch size).

Our ContGCN model is designed based on the all-token-any-document paradigm. We define all tokens as the vocabulary used for PLM tokenizers, allowing unseen words to be tokenized into sub-words that are already present in the vocabulary. 
When a new batch of data is fed into the model, the adjacency matrix and node embedding will be dynamically updated by the adjacency matrix generator and node encoder, respectively.

\subsection{Adjacency Matrix Generator}


As illustrated in Figure~\ref{fig:ATAD}, the adjacency matrix consists of three matrices: a token-token matrix $\am^{(1)} \mmSqr{\nVoc}$, a document-token matrix $\am^{(2)} \mm{\nBat}{\nVoc}$, and a document-document identity matrix $\am^{(3)} \mmSqr{\nBat}$.

The token-token matrix $\am^{(1)}$ is learned from the token occurrence knowledge of the corpus and is \textit{phase-fixed}. This means that it will be refined when the model enters a new training or testing phase with the emergence of new corpora and token co-occurrence knowledge.
The document-token matrix $\am^{(2)}$ is actively calculated based on the current batch of data and is used to update the adjacency matrix dynamically.
Finally, the inner-document identity matrix $\am^{(3)}$ ensures that each document is not influenced by other samples during model learning or reasoning.

The Occurrence Memory Module (OMM) is a module that incrementally records historical statistics, which includes a document counter $\nSen \in \mathbb{Z}^1$ that keeps track of the number of sentences, a token occurrence counter $\mathbf{c} \in \mathbb{Z}^{\nVoc}$ that records the number of sentences in which a token appears, and a token co-occurrence counter $\mathbf{C} \in \mathbb{Z}^{\nVoc \times \nVoc}$ that records the number of times two tokens appear in the same sentence. The OMM captures global non-consecutive semantic information and offers the following benefits: 1) it enables the dynamic calculation of the adjacency matrix for any batch of documents, and 2) it stores a large amount of previous general and domain-specific knowledge without the need for recalculation during updates. We develop a simple yet effective algorithm for updating the OMM, as described in Algorithm~\ref{algo:omm}. As shown in Figure~\ref{fig:overview}, the OMM is initialized with the Wikipedia corpus and updated by the training or testing data before model training or testing.
Thus, $\am^{(1)}$ will be phasely updated by PPMI (positive pointwise mutual information), which is defined as:
\begin{gather}
    \am^{(1)}_{i, j} = 
	\begin{cases}
		1, & \mbox{if } i = j, \\
		\max \left( \log \left( \nSen\frac{C_{i, j}}{c(i,:)c_{j}} \right), 0 \right), & \mbox{else.}
	\end{cases}
\end{gather}

To obtain the document-token correlation for each document $\mathbf{s} \in B$, we use TF-IDF (term frequency-inverse document frequency), which is calculated as:
\begin{gather}
    \am^{(2)}_{\mathbf{s}, t} = \frac{\mathtt{g}(\mathbf{s}, t)}{|\mathbf{s}|} \log  \frac{\nSen}{c_t + 1},
\end{gather}
where $\mathtt{g}(\mathbf{s}, t)$ represents the number of times token $t$ appears in document $\mathbf{s}$. 
The inner-document matrix $\am^{(3)}$ is an identity matrix, denoted as:
\begin{gather}
    \am^{(3)}_{i, j} = 
	\begin{cases}
		1, & \mbox{if } i = j, \\
		0, & \mbox{else.}
	\end{cases}
\end{gather}

Finally, the adjacency matrix $\am$ can be composed by:
\begin{gather}
    \am = \begin{pmatrix} \am^{(1)} & \am^{(2)\top} \\ \am^{(2)} & \am^{(3)} \end{pmatrix}.
\end{gather}

\begin{algorithm}[t]
\SetKwInOut{Input}{Input}
\SetKwInOut{Output}{Output}
\SetKwFunction{len}{len}
\SetKwFunction{set}{set}

\Input{A corpus or dataset $\mathcal{D}$, and OMM counters $s$, $\mathbf{c}$, and $\mathbf{C}$}
\BlankLine

Update document counter, i.e., $\nSen \leftarrow \nSen + \len(\mathcal{D})$;

\For{$\textnormal{each document in the corpus or dataset } \mathcal{D}$}{
    \For{$\textnormal{each sentence in the document }$}{
        Update the count of token $t_i$ if it appears in current sentence, i.e., $\mathbf{c}[t_i] \leftarrow \mathbf{c}[t_i] + 1$;
        
        \For{$\textnormal{each token pair } t_i \textnormal{ and } t_j \textnormal{ in the sentence}$}{
            \If{$t_i \neq t_j$}{
                Update the co-occurrence count, i.e., $\mathbf{C}[t_i][t_j] \leftarrow \mathbf{C}[t_i][t_j] + 1$;
            }
        }
    }
}
\Output{Updated OMM counters $s$, $\mathbf{c}$, and $\mathbf{C}$}
\caption{Continual OMM updating algorithm}\label{algo:omm}
\end{algorithm}

\subsection{Node Encoder}

The effectiveness of pretrained language models (PLMs) such as BERT~\cite{bert}, RoBERTa~\cite{liu2019roberta}, and XLNet~\cite{xlnet} for text modeling has been widely demonstrated in various scenarios. Therefore, we utilize a PLM as a document encoder to capture semantic information for each document $\mathbf{s} \in B$:
\begin{gather}
    \mathbf{E}_\ssp = \mathtt{PLM} (\mathbf{s}) \mm{\nLen}{\nDim}, \label{eq:plm}
\end{gather}
where $\nLen$ is the maximum document length for PLM, and each row of $\mathbf{E}$ is a token embedding. 
We append special $\mathtt{<PAD>}$ tokens to short documents to align their length with other documents in the batch, following BERT. We average the hidden states of the first and last Transformer layers following previous works~\cite{li-etal-2020-sentence,su2021whitening}.
We then perform an average pooling operation on $\mathbf{E}_\ssp$ to obtain the unified document embedding $\mathbf{d}^\ssp \in \mathbb{R}^{\nDim}$. 
Following BertGCN \cite{bertgcn}, we form a batch-wise node embedding $\mathbf{X}^{n} \mmDim{(\nVoc+\nBat)}$:
\begin{gather}
    \mathbf{X}^n = \left(\mathbf{0}, \cdots, \mathbf{0}, \mathbf{d}^{(\mathbf{s}_1)}, \cdots, \mathbf{d}^{(\mathbf{s}_{\nBat})} \right)^\top,
\end{gather}
where $\mathbf{d}^\sj$ is the embedding of document $j$ in the current batch $B$. 
However, the document embeddings will cause interference among one another when doing node message passing by GCN. 
To avoid interference, for each document $j$
in the batch, we form a sample-specific node embedding $\mathbf{X}^p_\sj \mmDim{(\nVoc+\nBat)}$ by:
\begin{gather}
    \mathbf{X}^p_\sj = \left(\mathbf{M}_\sj, \mathbf{0}, \cdots, \mathbf{0}, \mathbf{d}^\sj, \mathbf{0}, \cdots, \mathbf{0} \right)^\top,
\end{gather}
where $\mathbf{M}_\sj \mm{\nVoc}{\nDim}$ is a sample-specific token embedding matrix of document $j$:
\begin{gather}
    \mathbf{M}_\sj(i,:) = 
	\begin{cases}
		\mathbf{E}_\sj(k,:), & \mbox{if token } i \mbox{ of the vocabulary} \\ & \mbox{is the }k \mbox{-th token in } \mathbf{s}_j, \\
		\mathbf{0}, & \mbox{otherwise.}
	\end{cases} \label{eq:m}
\end{gather}

We refer to $\mathbf{X}^{n}$ as the \textit{jammed} node embeddings for all documents in the current batch, and $\mathbf{X}^p_\ssp$ as the \textit{unjammed} node embedding of a single document $\mathbf{s}$.

\subsection{GCN Encoder}

Once the adjacency matrix ($\am$) and node embeddings (unjammed $\mathbf{X}^p_\ssp$ and jammed $\mathbf{X}^n$) are generated, the GCN encoder is applied to obtain graph-enhanced node representations.
We denote the GCN-encoded unjammed and jammed node representations as $\bar{\mathbf{X}}^p_\ssp$ and $\bar{\mathbf{X}}^n$, respectively.
Next, we extract the document embeddings from $\bar{\mathbf{X}}^p_\sj (\forall \mathbf{s}_j\in B)$  to form $\mathbf{Z} \mmDim{\nBat}$ (see Figure~\ref{fig:overview}), i.e., 
\begin{equation}
    \mathbf{Z}(j,:) = \bar{\mathbf{X}}^p_\sj(j+\nVoc,:).
\end{equation}
$\mathbf{Z}$ will be used in the classification task. We then extract the document embeddings from $\bar{\mathbf{X}}^p_\sj$ to form $\mathbf{Z}^p_\sj \mmDim{\nBat}$ and those from $\bar{\mathbf{X}}^n$ to form $\mathbf{Z}^n \mmDim{\nBat}$ by:
\begin{gather}
    \mathbf{Z}^p_\sj(i,:) = \bar{\mathbf{X}}^p_\sj(i+\nVoc,:)~~ \text{and}\\
    \mathbf{Z}^n(i,:) = \bar{\mathbf{X}}^{n}(i+\nVoc,:),
\end{gather}
as illustrated in Figure~\ref{fig:overview}.
$\mathbf{Z}^p_\sj $ and $\mathbf{Z}^n $ will be used in the anti-interference contrastive task.

\subsection{Training Objectives}


To train the model, we employ two tasks: a document classification task and an anti-interference contrastive task.

The document classification task 
utilizes a multi-layer perceptron (MLP) classifier $f: \mathbb{R}^\nDim \rightarrow \mathbb{R}^\nCls$ ($\nCls$ is the number of document classes) with a softmax activation function to infer the probability distribution over all classes. The loss function can be defined as:
\begin{gather}
    \mathcal{L}_{\text{cls}} = -\frac{1}{\nBat} \sum^\nBat_{j=1} \log \left(f\left( \mathbf{Z}(j,:) \right)_{l_j}\right),
\end{gather}
where $l_j$ is the class label of the $j$-th document.

For the anti-interference contrastive task, the goal is to learn a representation space where semantically similar documents are closer to each other and dissimilar documents are farther apart. Hence, in the contrastive task, we enforce the GCN encoder to learn a mapping such that the jammed embedding of document $j$ (i.e., $\mathbf{Z}^n(j,:)$) is close to its unjammed embdding (i.e., $\mathbf{Z}^p_\sj(j,:)$) while distant from the embeddings of other documents (i.e., $\mathbf{Z}^p_\sj(k,:)$, $k \neq j$) in the batch. 
The loss function can be calculated by:
\begin{gather}
    \mathcal{L}_{\text{aic}} = -\frac{1}{\nBat} \sum^\nBat_{j=1} \log \left( \mathbf{y}_\sj(j) \right), \text{where}\label{eq:contrastive-loss}\\
    \mathbf{y}_\sj = \mathtt{softmax} \left(\mathbf{Z}^{p}_\sj \left(\mathbf{Z}^{n}(j,:)\right)^\top \right) \in \mathbb{R}^\nBat.
\end{gather}
The anti-interference contrastive task helps to learn representations robust to the interference between documents.

The overall loss function is a combination of the classification and contrastive tasks with a balancing parameter $\lambda$, denoted as:
\begin{equation}\label{eq:overall-loss}
    \mathcal{L} = \mathcal{L}_{\text{cls}} + \lambda \mathcal{L}_{\text{aic}}.
\end{equation}




\textbf{Label-free Updating Mechanism (LUM).} The occurrence memory module and the anti-interference contrastive task enables to continually update our ContGCN model with incoming unlabeled text data during inference. Hence, we name it label-free updating mechanism (LUM).


\subsection{Model Training and Update}

The training and updating of the ContGCN model involves three stages.

\textbf{Stage 1: Before training.} Prior to training, we perform post-pretraining on the pre-trained language model by using the classification task on the PLM-enhanced document embeddings. This pre-training helps to speed up the convergence of the model during the training process.

\textbf{Stage 2: During training.} During the training process, we use the multi-task training objective (Eq.~\ref{eq:overall-loss}) to train the ContGCN model. 

\textbf{Stage 3: During inference.} When new test data is available, we first update the occurrence memory module using Algorithm~\ref{algo:omm}. We then finetune the ContGCN model using the auxiliary anti-interference contrastive task (Eq.~\ref{eq:contrastive-loss}) to improve model performance during inference.


\section{Experiments \label{sec:experiments}}


\newcommand{\contplm}[1]{{ContGCN$_\texttt{#1}$}}
\newcommand{\contbert}{\contplm{BERT}}
\newcommand{\contxl}{\contplm{XLNet}}
\newcommand{\contro}{\contplm{RoBERTa}}

\subsection{Experimental Setups}
\subsubsection{Datasets.}
As described in~\cite{bertgcn}, we have performed experiments on five text classification datasets which are commonly used in real-world applications: 20-Newsgroups (20NG), Ohsumed, R52 Reuters, R8 Reuters, and Movie Review (MR) datasets. Table~\ref{tab:statistics} presents the summarized statistics of these datasets. We randomly chose 10\% of the training set for validation purposes for all datasets.

\subsubsection{Baselines and Variants of Our Method.} To validate the effectiveness of our proposed ContGCN model, we compare it with three types of state-of-the-art models:: 1) traditional GCN-based models that do not utilize pretrained general semantic knowledge, including TextGCN~\cite{yao2019graph} and TensorGCN~\cite{liu2020tensor};
2) transformer-based PLMs, including BERT~\cite{bert}, RoBERTa~\cite{liu2019roberta} and XLNet~\cite{yang2019xlnet};
3) models that combine GCN with PLM, including TG-Transformer~\cite{zhang2020text}, BertGCN~\cite{bertgcn} and RoBERTaGCN~\cite{bertgcn}. As our ContGCN can be plugged by different PLMs, we adopt BERT, XLNet, and RoBERTa as alternatives, denoted as \contbert, \contxl, and \contro.

\subsubsection{Implementation Details.}
We adopt the Adam optimizer~\cite{kingma2014adam} to train the network of our ContGCN model and the baseline models, which are consistent in the following parameters. The following parameters are kept consistent across all models: the number of graph convolutional layers, if applicable, is set to 3; the embedding dimension is set to 768; and the batch size is set to 64. In the post-pretraining phase, we set the learning rate for PLM to 1e-4. 
During training, we set different learning rates for PLM and other randomly initialized parameters including those of the GCN network, following~\citet{bertgcn}. Precisely, we set 1e-5 to finetune RoBERTa and BERT, 5e-6 to finetune XLNet, and 5e-4 to other parameters. 
We average the results of 10 runs as the final evaluation results.

\begin{table}[t]
\centering 

\resizebox{1.0\linewidth}{!}{
\begin{tabular}{l|ccccc}
\toprule
\textbf{Dataset} & \textbf{20NG} & \textbf{R8} & \textbf{R52} & \textbf{Ohsumed} & \textbf{MR} \\
\midrule
\# Docs & 18,846 & 7,674 & 9,100 & 7,400 & 10,662 \\
\# Training & 11,314 & 5,485 & 6,532 & 3,357 & 7,108 \\
\# Test & 7,532 & 2,189 & 2,568 & 4,043 & 3,554 \\
\# Classes & 20 & 8 & 52 & 23 & 2 \\
Avg. Length & 221 & 66 & 70 & 136 & 20 \\
\bottomrule
\end{tabular}
}

\caption{\label{tab:statistics} Dataset statistics.}
\end{table}

\begin{table}[t]
\centering 

\resizebox{1.0\linewidth}{!}{
\begin{tabular}{l|ccccc}
\toprule
\textbf{Models} & \textbf{20NG} & \textbf{R8} & \textbf{R52} &  \textbf{Ohsumed} & \textbf{MR} \\
\midrule
TextGCN & 86.3 & 97.1 & 93.6 & 68.4 & 76.7 \\
TensorGCN & 87.7 & 98.0 & 95.0 & 70.1 & 77.9 \\
\midrule
BERT & 85.3 & 97.8 & 96.4 & 70.5 & 85.7 \\
RoBERTa & 83.8 & 97.8 & 96.2 & 70.7 & 89.4 \\
XLNet & 85.1 & 98.0 & \underline{96.6} & 70.7 & 87.2 \\
\midrule
TG-Transformer & - & 98.1 & 95.2 & 70.4 & - \\
BertGCN & 89.3 & 98.1 & \underline{96.6} & \underline{72.8} & 86.0 \\
RoBERTaGCN & \underline{89.5} & \underline{98.2} & 96.1 & \underline{72.8} & \underline{89.7} \\
\midrule
\contbert & 89.4 & 98.3 & 96.9 & 73.1 & 86.4 \\
\contxl & 89.7 & 98.5 & \textbf{97.0} & 73.1 & 88.7 \\
\contro & \textbf{90.1} & \textbf{98.6} & 96.6 & \textbf{73.4} & \textbf{91.3} \\
\bottomrule
\end{tabular}
}
		
\caption{\label{tab:big-table} Comparison of ContGCN with state-of-the-art models in offline evaluation. The best results are in boldface, and the second best results are underlined.}
\end{table}

\subsection{Offline Evaluation}

We conduct an offline evaluation of our ContGCN model and state-of-the-art baselines on five datasets.
Table~\ref{tab:big-table} presents the overall performance of all methods, and the following observations can be made. \textbf{First}, PLM-only methods mostly outperform GCN-only methods due to their pre-learned semantic knowledge. However, GCN-only methods can build more document-token edges for better semantic comprehension on datasets with long documents, such as 20NG, but they struggle on datasets with short documents, such as MR. \textbf{Second}, PLM-empowered GCN methods combine the strengths of both PLMs and GCNs and outperform both PLM-only and GCN-only methods. \textbf{Third}, our ContGCN model achieves state-of-the-art performance on five datasets, thanks to 1) the proposed all-token-any-document paradigm that leverages general semantic knowledge from a large Wikipedia corpus and 2) the proposed contrastive learning objective that reduces inter-document interference. Notably, our \contro{} achieves the best performance on four datasets.

\begin{table}[t]
\centering 

\begin{tabular}{l|ccc}
\toprule
\textbf{Models} & \textbf{20NG} & \textbf{R8} & \textbf{Ohsumed} \\
\midrule
\contro & 90.1 & 98.6 & 73.4 \\
\quad w/o Wikipedia Init & 89.9 & 98.2 & 73.1 \\
\quad w/o OMM Updating & 89.6 & 98.3 & 73.0 \\
\quad w/o Contrastive Loss & 89.7 & 98.5 & 73.2 \\
\midrule
\contxl & 89.7 & 98.5 & 73.1 \\
\quad w/o Wikipedia Init & 89.8 & 98.3 & 72.8\\
\quad w/o OMM Updating & 89.4 & 98.2 & 72.7 \\
\quad w/o Contrastive Loss & 89.5 & 98.2 & 73.0\\
\bottomrule
\end{tabular}
\caption{\label{tab:ablation} Influence of Wikipedia initialization, OMM updating, and the anti-interference contrastive task.}
\end{table}

\begin{table}[t]
\centering 

\begin{tabular}{l|cccccc}
\toprule
\textbf{Variants} & \textbf{1/6} & \textbf{2/6} & \textbf{3/6} & \textbf{4/6} & \textbf{5/6} & \textbf{6/6} \\
\midrule
ContGCN$^*$ & 86.4 & 87.3 & 88.1 & 88.6 & 89.0 & 89.6 \\
\midrule
ContGCN & 86.3 & 87.1 & 87.8 & 88.2 & 88.7 & 89.1 \\
ContGCN$^\alpha$ & 86.1 & 86.9 & 87.5 & 87.9 & 88.3 & 88.7 \\
ContGCN$^\beta$ & 86.0 & 86.2 & 86.4 & 86.6 & 86.9 & 87.1 \\
\bottomrule
\end{tabular}
\caption{\label{tab:ablation-online} Comparisons of variants of \contro{} in the online learning scenario on the 20NG dataset. ContGCN$^*$ is retrained from scratch in each session with all previously seen data. ContGCN$^\alpha$ is updated without the contrastive loss. ContGCN$^\beta$ is updated without LUM.}
\end{table}

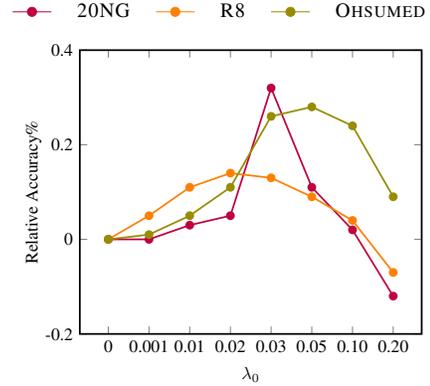
\begin{figure}
\centering

\begin{tabular}{c}
\resizebox{0.7\linewidth}{!}{
\begin{tikzpicture}
    \begin{customlegend}[
        legend columns=3,
        legend style={
            align=left,
            draw=none,
            column sep=2ex
        },
        legend entries={
            \textsc{\small{20NG}},
            \textsc{\small{R8}},
            \textsc{\small{Ohsumed}},
        }]
        \addlegendimage{purple,mark=*,solid,line legend}
        \addlegendimage{orange,mark=*,solid,line legend}
        \addlegendimage{olive,mark=*,solid,line legend}
        \end{customlegend}
\end{tikzpicture}

}
\\
\resizebox{0.65\linewidth}{!}{
    \begin{tikzpicture}
    \begin{axis}[
        xtick={1,2,3,4,5,6,7,8},
        xticklabels={0, 0.001, 0.01, 0.02, 0.03, 0.05, 0.10, 0.20},
        xlabel=$\lambda_0$,
        line width=0.35mm,
        yticklabels={-0.2, -0.2, 0, 0.2, 0.4},
        ymin=-0.2,
        ymax=0.4,
        ylabel=Relative Accuracy\%,
        ylabel near ticks,
        ylabel shift={-5pt},
    ]

        \addplot[
            purple,
            mark=*,
        ]
            coordinates{
                (1, 0) (2, 0.00) (3, 0.03) (4, 0.05) (5, 0.32) (6, 0.11) (7, 0.02) (8, -0.12)  
            }; 
            
        \addplot[
            orange,
            mark=*,
        ]
            coordinates{
                (1, 0) (2, 0.05) (3, 0.11) (4, 0.14) (5, 0.13) (6, 0.09) (7, 0.04) (8, -0.07) 
            }; 
            
        \addplot[
            olive,
            mark=*,
        ]
            coordinates{
                (1, 0) (2, 0.01) (3, 0.05) (4, 0.11) (5, 0.26) (6, 0.28) (7, 0.24) (8, 0.09) 
            }; 

    \end{axis}

\end{tikzpicture}

}
\end{tabular}

\caption{\label{fig:lambda} Influence of the parameter $\lambda$ that weights the anti-interference contrastive loss.
\emph{Relative accuracy (\%)} means the difference between the accuracy achieved with $\lambda = \lambda_0$ and that achieved with $\lambda = 0$.}
\end{figure}

\subsection{Ablation Study}

First, we study the effect of different components of ContGCN, including Wikipedia initialization, OMM updating, and anti-interference contrastive task on offline performance.
Based on the results from Table~\ref{tab:ablation}, we can conclude the following. \textbf{First}, on the 20NG dataset, we observed that Wikipedia initialization is less effective, which is likely because the lengthy documents already contain sufficient non-consecutive knowledge during OMM updating. Apart from this, removing any of these components leads to a decline in performance for both \contro{} and \contxl{} on three datasets, confirming their effectiveness. \textbf{Second}, models without OMM updating show the worst performance, indicating the importance of non-consecutive semantic information in training and testing.

Next, we study the updating strategy of ContGCN in the online learning scenario. As shown in Table~\ref{tab:ablation-online}, we can make the following observations. \textbf{First}, by comparing ContGCN with ContGCN$^\alpha$ and ContGCN$^\beta$, it can be seen that both OMM updating and the contrastive loss are effective.
\textbf{Second}, compared with retraining from scratch, i.e., ContGCN$^*$, ContGCN requires less computational resources and time to update yet still achieves competitive performance.

\subsection{Impact of Anti-interference Contrastive Learning}

Here, we study the balancing parameter $\lambda$ which weights the auxiliary anti-interference contrastive loss. 
We conduct experiments on the 20NG, R8 and Ohsumed datasets with \contro{} model, varying $\lambda$ in \{0.001, 0.01, 0.02, 0.03, 0.05, 0.10, 0.20\}.
As demonstrated in Figure~\ref{fig:lambda}, we can make the following observations. \textbf{First}, 
the reliance on the auxiliary task varies for different datasets. 
Specifically, the model achieves the best performance on the 20NG, R8, and Ohsumed datasets when $\lambda$ is set to 0.03, 0.02, and 0.05, respectively. \textbf{Second}, for each dataset, as $\lambda$ increases, the performance first increases and then decreases. Hence, it is essential to select a good $\lambda$.

\subsection{Online Evaluation}


\subsubsection{Fixed Testing Data.} 
\begin{figure}
\centering
\setlength\tabcolsep{2pt}
\begin{tabular}{cc}
\multicolumn{2}{c}{
    \resizebox{1.0\linewidth}{!}{
\begin{tikzpicture}
    \begin{customlegend}[
        legend columns=3,
        legend style={
            align=left,
            draw=none,
            column sep=2ex
        },
        legend entries={
            \textsc{\small{TextGCN}},
            \textsc{\small{RoBERTa}},
            \textsc{\small{RoBERTaGCN}},
        }]
        \addlegendimage{purple,mark=x,dotted,line legend}
        \addlegendimage{orange,mark=x,solid,line legend}
        \addlegendimage{olive,mark=x,dotted,line legend}
        \end{customlegend}
\end{tikzpicture}
    }
} \\
\multicolumn{2}{c}{
    \resizebox{0.95\linewidth}{!}{
\begin{tikzpicture}
    \begin{customlegend}[
        legend columns=2,
        legend style={
            align=left,
            draw=none,
            column sep=2ex
        },
        legend entries={
            \textsc{\small{RoBERTaGCN$_\texttt{scratch}$}},
            \textsc{\small{\contro}},
        }]
        \addlegendimage{olive,mark=x,solid,line legend}
        \addlegendimage{violet,mark=x,solid,line legend}
        \end{customlegend}
\end{tikzpicture}
    }
} \\
\begin{subfigure}{0.23\textwidth}
    \resizebox{1.0\linewidth}{!}{
\begin{tikzpicture}
    \begin{axis}[
        xtick={1,2,3,4,5,6,7},
        xticklabels={0/6, 1/6, 2/6, 3/6, 4/6, 5/6, 6/6},
        line width=0.35mm,
        ytick={78, 82, 86, 90},
        ymin=78,
        ymax=90,
        ylabel=Accuracy,
        ylabel near ticks,
        ylabel shift={-5pt},
    ]

        \addplot[
            purple,
            dotted,
            mark=x,
        ]
            coordinates{
                (1, 80.5) (2, 80.5) (3, 80.5) (4, 80.5) (5, 80.5) (6, 80.5) (7, 80.5) 
            }; 
            
        \addplot[
            orange,
            mark=x,
        ]
            coordinates{
                (1, 81.2) (2, 81.5) (3, 81.3) (4, 81.1) (5, 81.9) (6, 81.8) (7, 82.1)
            }; 
            
        \addplot[
            olive,
            dotted,
            mark=x,
        ]
            coordinates{
                (1, 84.8) (2, 84.8) (3, 84.8) (4, 84.8) (5, 84.8) (6, 84.8) (7, 84.8)
            }; 
            
        \addplot[
            olive,
            mark=x,
        ]
            coordinates{
                (1, 84.8) (2, 85.2) (3, 85.5) (4, 86.1) (5, 86.6) (6, 87.1) (7, 87.9)
            }; 
            
        \addplot[
            violet,
            mark=x,
        ]
            coordinates{
                (1, 85.7) (2, 86.3) (3, 87.1) (4, 87.8) (5, 88.2) (6, 88.7) (7, 89.1)
            }; 
    \end{axis}

\end{tikzpicture}

    }
    \caption{\label{fig:online-quality}Reasoning Quality}
\end{subfigure} & 
\begin{subfigure}{0.23\textwidth}
    \resizebox{1.0\linewidth}{!}{
\begin{tikzpicture}
\begin{axis}[
    xtick={1,2,3,4,5,6},
    xticklabels={1/6, 2/6, 3/6, 4/6, 5/6, 6/6},
    line width=0.35mm,
    ymin=0.7,
    ymax=1.5,
    ylabel=Finetuning Time Ratio,
    ylabel near ticks,
    ylabel shift={-5pt},
]
        
    \addplot[
        orange,
        mark=x,
    ]
        coordinates{
            (1, 1.00) (2, 0.78) (3, 0.89) (4, 1.03) (5, 0.82) (6, 0.92)
        }; 
        
    \addplot[
        olive,
        mark=x,
    ]
        coordinates{
            (1, 1.00) (2, 1.07) (3, 1.12) (4, 1.15) (5, 1.24) (6, 1.34) 
        }; 
        
    \addplot[
        violet,
        mark=x,
    ]
        coordinates{
            (1, 1.00) (2, 0.95) (3, 0.94) (4, 0.98) (5, 0.95) (6, 1.02) 
        }; 

\end{axis}

\end{tikzpicture}

    }
    \caption{\label{fig:online-efficiency}Finetuning Efficiency}
\end{subfigure}
\end{tabular}

\caption{\label{fig:online} Comparison between our ContGCN model and baselines in an online learning scenario. We divide the 20NG dataset into training, testing, and updating sets by the ratio of 2:2:6. We trained each model with the training set to learn an initial version. Then, we divided the updating set into six equal parts and gradually fed each part to the model for fine-tuning. The \textit{finetuning time ratio} in (b) is calculated by the finetuning time of the current session over that of the first session. For each training or updating session, we used 10\% of the training set as the validation set. }
\end{figure}
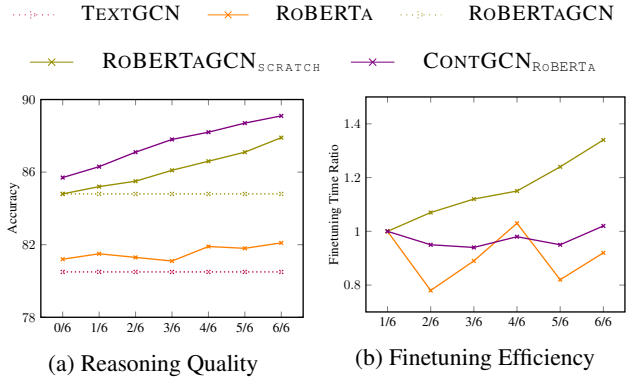

Figure~\ref{fig:online} illustrates the performance of our model and baselines in an online learning scenario where the training/updating data is incremental and the testing data is fixed. Based on the results, we can draw the following observations: 
\textbf{First}, GCN-based methods such as TextGCN and RoBERTaGCN cannot be updated and are incapable of reasoning about unobserved data as they construct fixed graphs based on the original corpus. Therefore, their performance remains constant over time, as shown by the dotted lines in Figure~\ref{fig:online-quality}.
\textbf{Second}, as the updating data increases, the performance of all updatable models improves.
\textbf{Third}, our proposed ContGCN method outperforms all baselines in all sessions.
\textbf{Fourth}, due to the limitations of STSD-based GCN methods, we introduce a RoBERTaGCN$_\texttt{scratch}$ model that retrains from scratch in each session with all previously seen data. However, this model still falls short compared to \contro due to the utilization of Wikipedia Initialization and the unjammed node embedding in \contro. Moreover, as illustrated in Figure~\ref{fig:online-efficiency}, the updating time of RoBERTaGCN$_\texttt{scratch}$ increases almost linearly w.r.t. data size, making it unsuitable for online learning. The finetuning time ratios of ContGCN and RoBERTa are closer to or less than 1, indicating that for these models, each session takes up a comparable amount of time. 

\subsubsection{Fixed Training Data.} We have deployed ContGCN on Huawei public opinion analysis system, under the scenario that the training data is fixed and the updating/testing data is incremental. We use an optimized variant of RoBERTa~\cite{liu2019roberta} -- RoBERTa$_\texttt{wwm-ext}$~\cite{xu2021roberta}, tailored for Chinese text classification, which will still be referred to as RoBERTa below and in Table~\ref{tab:industrial}. 
We implement RoBERTaGCN by plugging RoBERTa into BertGCN~\cite{bertgcn}, which cannot infer unobserved documents due to the STSD scheme and hence the results are unavailable in the following months.
We then implement ContGCN based on RoBERTa and the proposed ATAD scheme, i.e., \contro{}. After RoBERTa and \contro{} were trained offline initially (i.e., in the 0th month), we deploy them online for comparison. As illustrated in Table~\ref{tab:industrial}, our \contro{} has achieved 5.94\%, 6.57\%, 7.98\%, and 8.86\% gains in accuracy over RoBERTa in the 0th, 1st, 2nd, and 3rd month respectively. Besides, due to the \textit{distribution shift} of public opinions, the accuracy drops slightly over time. However, \contro{} still maintains higher performance than RoBERTa. 
Furthermore, by removing the label-free update mechanism (i.e., \contro{}$^\beta$), the performance drops significantly, which demonstrates the capability of our ContGCN model in continual learning.

\begin{table}[t]
\centering 

\begin{tabular}{l|cccc}
\toprule
\textbf{Models} & \textbf{0th} & \textbf{1st} & \textbf{2nd} & \textbf{3rd} \\
\midrule
RoBERTaGCN & 91.7 & N/A & N/A & N/A \\
\midrule
RoBERTa & 87.6 & 86.8 & 85.2 & 83.5\\
ContGCN$_\texttt{RoBERTa}^\beta$  & \textbf{92.8} & 90.3 & 89.9 & 88.2 \\
\contro & \textbf{92.8} & \textbf{92.5} & \textbf{92.0} & \textbf{90.9}\\
\bottomrule
\end{tabular}
\caption{\label{tab:industrial} Comparison of our ContGCN model with RoBERTa in an industrial online learning scenario. All models are first trained offline (in the 0th month) with a labeled dataset. After deployed, \contro{} performs online learning with LUM. ContGCN$_\texttt{RoBERTa}^\beta$ is a static network with parameters fixed after the initial training.
}
\end{table}

\section{Related Work \label{sec:rw}}

\paragraph{Graph Convolutional Network.}
Graph Convolutional Networks~\cite{gcn} (GCNs) have become increasingly popular in recent years due to their ability to capture the structural relations among data~\cite{hamilton2017inductive,li2018deeper}. They can learn representations of graph data by propagating information between nodes in the graph. The popularity of GCNs can be attributed to their versatility and effectiveness in various applications~\cite{zhang2019graph}, including image classification~\cite{hong2020graph}, video understanding~\cite{huang2020location}, social recommendation~\cite{fan2019graph}, and text classification~\cite{yao2019graph,bertgcn}.

\paragraph{Text Classification.} Text classification is a critical and fundamental task in the field of natural language processing. Early studies~\cite{jacovi-etal-2018-understanding,JITEKI15021} utilize word embedding methods~\cite{mikolov2013distributed,pennington2014glove} or apply traditional models~\cite{zhang2015character,lai2015recurrent} such as convolutional neural networks~\cite{cnn} or recurrent neural networks~\cite{rumelhart1985learning} to learn textual knowledge. In recent years, Transformer-based pretrained language models such as BERT~\cite{bert} have been introduced~\cite{sun2019fine} for text classification due to its strong ability of semantic comprehension. However, these models do not effectively utilize global semantic information such as token co-occurrence in a corpus.

\paragraph{GCN-based Text Classification.} 
GCNs have been gaining attention in text classification, owing to their ability to model non-structured data and capture global dependencies, such as high-order neighborhood information~\cite{yao2019graph,bertgcn}. Unlike some GCN-based methods~\cite{li2019label,xie-etal-2021-inductive,li2021dimensionwise} that construct homogeneous document graphs, we follow \citet{yao2019graph} to construct a document-token graph to better capture semantic relations. In contrast to previous methods that typically construct fixed graphs and are limited in their ability to reason about unobserved documents, our proposed ContGCN model employs an all-token-any-document paradigm, enabling making inferences on new data in online systems. 

\section{Conclusion}

To deploy GCN-based text classification methods to online industrial systems, we propose a ContGCN model with a novel all-token-any-document paradigm and a label-free updating mechanism, which endow the model the ability to infer unobserved documents and enable to continually update the model during inference time. To our knowledge, this is the first attempt to use GCN for online text classification. Extensive online and offline evaluations validate the effectiveness of ContGCN, which achieves favorable performance compared with various state-of-the-art methods.

\section*{Acknowledgments}

The authors would like to thank Qimai Li for valuable discussion and the anonymous reviewers for their helpful comments. Q. Liu and X. Wu were supported by GRF No.15222220 funded by the UGC and ITS/359/21FP funded by the ITC of Hong Kong.

\bibliography{ContGCN}

\end{document}